\documentclass[a4paper, 10 pt, conference]{IEEEtran} 
\usepackage{cite}
\usepackage{footnote}
\usepackage[ruled,vlined]{algorithm2e}
\usepackage{comment}
\usepackage{multirow}
\usepackage{amssymb}
\usepackage[hyphens]{url}
\usepackage{color,soul}
\usepackage{subfig}
\usepackage{amsmath}
\usepackage{threeparttable}
\usepackage{balance}
\usepackage{float}
\usepackage{soul}
\usepackage{makecell}
\usepackage{rotating}
\usepackage[table]{xcolor}

\begin{document}

\title{Mechanistic Interpretation of Machine Learning Inference: A Fuzzy Feature Importance Fusion Approach}
\author{Divish Rengasamy$^{1}$, Jimiama M. Mase$^{1}$, Mercedes Torres Torres, Benjamin Rothwell\\
 David A. Winkler,  Grazziela P. Figueredo}

\maketitle

\begin{abstract}

With the widespread use of machine learning to support decision-making, it is increasingly important to verify and understand the reasons why a particular output is produced. Although post-training feature importance approaches assist this interpretation, there is an overall lack of consensus regarding how feature importance should be quantified, making explanations of model predictions unreliable. In addition, many of these explanations depend on the specific machine learning approach employed and on the subset of data used when calculating feature importance. A possible solution to improve the reliability of explanations is to combine results from multiple feature importance quantifiers from different machine learning approaches coupled with re-sampling. Current state-of-the-art ensemble feature importance fusion uses crisp techniques to fuse results from different approaches. There is, however, significant loss of information as these approaches are not context-aware and reduce several quantifiers to a single crisp output. More importantly, their representation of ``importance'' as coefficients is misleading and incomprehensible to end-users and decision makers. Here we show how the use of fuzzy data fusion methods can overcome some of the important limitations of crisp fusion methods. 

\end{abstract}


\section{Introduction}

The ubiquitous application of machine learning (ML) to a rapidly increasing number of important applications has resulted in sophisticated and effective solutions with little to no human intervention. As the complexity of these ML models increases, however, understanding why and how decisions are made is important yet challenging.  For safety-critical systems, how ML outputs are generated is especially important for model verification, regulatory compliance~\cite{Reddy2019}, elucidation of ethical concerns, trustworthiness~\cite{GILLE2020100001}, and model diagnostics~\cite{rengasamy2020deep, rengasamy2020towards,mase2020capturing,arrieta2020explainable,agrawal2019towards}. One way to explain ML models outputs is by post-training analysis, such as feature importance (FI) analysis. FI estimates the contribution of each feature (independent parameter) to the model output~\cite{hooker2018evaluating,rengasamy2020towards}. Using FI to understand how decisions are made aids the establishment of true causality between important data attributes and outcomes in model inference~\cite{arrieta2020explainable,goudet2018learning}. In interdisciplinary contexts, where ML researchers work with other domain experts, robust identification of important variables and their influence on ML model outputs facilitates expert validation of the domain-specific processes being modelled~\cite{belle2017logic}. The elucidation of mechanistic processes linking the features to the outputs can validate results and lead to confidence in the model~\cite{lane2005explainable,lipton2018mythos,RyanMark2020}, acceptance, and reassurance of a fair intelligent tool being implemented~\cite{theodorou2017designing,chouldechova2017fair}; or reveal biases in data or model training and establish new requirements for model refinement. 

The wide availability of ML models and diversity of FI techniques complicates model selection, FI approaches used, reliability of interpretation of results from a given ML method coupled to a FI approach, and data representativeness. Different ML models may generate different FI values due to variations in their learning algorithms. Similarly, FI techniques may produce different importances with different ML algorithms or training data. For example, models that can adequately map the input to output relationships using linear functions can generate clear FIs, while relationships that are significantly nonlinear, FI can be a local, context-dependent property of the response surface. To add to the complication, some FI methods are model-agnostic, while others are model-specific.

To address these uncertainties in FI, Rengasamy \textit{et al.}~\cite{rengasamy2020towards} proposed an ensemble feature importance (EFI) method. Here, multiple FI approaches are applied to a set of ML models and their crisp importance values are combined to produce a final importance for each feature. This resulted in more robust and accurate analyses (15\% less FI error) of synthetic data sets compared to the use of a single ML and FI. However, their study had sampling limitations (train-test split) that prevented a comprehensive exploration of the feature space and feature causality. Furthermore, they ignored the fact that the ensemble FI values contain significantly more information than what is captured by crisp values. Most importantly, representation of FI as crisp coefficients can be misleading and incomprehensible for decision making.

To overcome those limitations, we propose a novel Fuzzy EFI (FEFI) method that captures and models the variance of different ML methods and FI techniques used to generate FI and data space representation. It combines crisp importance values for each feature from different ML models using Fuzzy Sets (FSs)~\cite{zadeh1965fuzzy}, with rules couched in simple linguistic terms. FEFI outputs three sets of membership functions (MFs)~\cite{zadeh1965fuzzy} determining: 

\begin{enumerate}
    \item  for each ML approach, which feature has low, moderate and high importance after training;  
    \item for each feature, its low, moderate and high importance relative to the other features in the data for each ML approach; 
    \item for each feature, its low, moderate, high importance  after training and after combining with importances from multiple ML approaches.
\end{enumerate}

The sets of MFs are obtained from the FI coefficients generated by applying multiple (ML, FI) combinations to the dataset using resampling (e.g., by cross-validation). This allows the variance between the ML models, FI techniques and data to be modelled as FSs, where `low’, `moderate’ and `high’ are the linguistic labels representing the relative importance of a feature. The Wang-Mendel~\cite{wangMendel} approach is employed to learn the rules that map the set of FI values to the final description of importance. Lastly, the Mamdani inference approach~\cite{mamdani1975experiment} is used to combine the FI terms generated by the MFs, using rules generated by Wang-Mendel, to produce a final description of feature importance that is easy to interpret for decision-making.

\section{Background}
\label{Background}

\subsection{Problem Statement}

Interpretation of feature importance in linear models is relatively straightforward. The regression coefficients for each feature in a multiple linear regression model (or generalised linear model) describe the magnitude and sign of their effect on the dependent variable. The importance of each feature is a global and invariant property across the model response surface (which consists of a set of linear hyperplanes). However, for ML models that exhibit significant non-linearity, the response surface is curved. This complicates the interpretation of feature importance because these are now local rather than global properties.

Estimating the importance of features in ML predictive analytics is currently very unreliable. Different ML models, FI techniques, and subsets of data generate different importance coefficients, often with diverse magnitudes, for the same features ~\cite{rengasamy2020towards}. These uncertainties in identifying the contribution of features to ML outputs are due to:

\begin{enumerate}

\item For nonlinear response surfaces (the multidimensional surfaces encoded by trained ML models), the FI depends on which part of the response surface is interrogated. The variation in feature gradients from one position to another is a result of the nonlinear behaviour of real-world systems. This variation means that for nonlinear models, the FI is not fixed, but is context dependent. Sensible choices for the part of the response surfaces at which to evaluate FI are typical local or global minima or maxima in the response variable, depending on what type of optimum is relevant.

\item The variance due to the characteristics of the learning process of ML models. For example, random forests~\cite{breiman2001random} combine the outputs of decision trees at the end of the training process while gradient boosting machines~\cite{friedman2001greedy} start combining at the start of the process. Although random forest and gradient boosting machines are tree-based methods that use impurity-based methods to calculate feature importance, the final feature importances for two ML methods will differ due to their variations in training algorithms.


\item Differences in how FI is calculated and interpreted by FI techniques. For instance, certain approaches, such as permutation importance (PI)~\cite{altmann2010permutation} investigate how each feature affects the model response individually. Here, PI shuffles the instance values for a particular feature, while maintaining the original values for the remaining features. The feature with shuffled instances is considered important if the model's performance decreases. Other approaches, however, investigate the importance of the feature both individually and in synergy with other subsets of features. For example, SHapley Addictive exPlanations (SHAP)~\cite{kumar2020problems} calculates the contribution of features for every possible combination of the feature set investigated. 

\item FI coefficients in general being calculated as the average (or weighted average) of the importance of a feature within a data sample. Information about the context (or data subspace) in which a feature has higher or lower importance during training is lost.

\end{enumerate}

\subsection{Related Work}
Earlier ensemble FI strategies for ML models focused on combining the importance of features from multiple decision trees. For example, Breiman~\cite{breiman2001random} used the Gini impurity metric across decision trees to calculate feature importance. Subsequently, De Bock~\textit{et al.}~\cite{de2012reconciling} extended the idea of feature-importance fusion from multiple weak learners to generalised additive models (GAM). Here, each weak learner undergoes permutation importance (PI) to calculate FI and the learners results are averaged. Zhai and Chen~\cite{zhai2018development} improved ensemble FI by using multiple ML models and gains in gini importance defining the final FI. Rengasamy~\textit{et al.}~\cite{rengasamy2020towards}, proposed a model agnostic ensemble FI framework to improve FI quantification using multiple models and multiple FI calculation methods. They studied several crisp fusion metrics such as mean, median, majority vote, rank correlation, combination with majority vote, and modified Thompson tau test. Majority voting produced the best overall results for several synthetic datasets. They showed that this ensemble of FI techniques, coupled with multiple ML approaches, produced more robust importance compared to single FI methods. This framework has been used to identify features relevant to the creep rate of additive manufactured materials~\cite{sanchez2021machine}. 


\subsection{Limitations of Existing Methods}
These single or multiple FI techniques suffer from several limitations. They have limited interpretability; the methods and outputs are difficult for domain experts to understand. For example, what does a final FI coefficient of 0.76 mean (is it a good high value, or a low value)? The value obtained also depends on the FI method adopted and those that are not model agnostic also depend on the specific ML model(s) employed. Use of crisp fusion can also result in significant loss of information as data subspaces and gradients are not adequately explored because of the way training/test sampling is conducted. Additionally, crisp ensembles like majority voting discard extreme values of FI which, for safety critical systems, should not be disregarded. Finally, there is a paucity of literature that explores data sampling for FI. Using solely training or test sets for FI leads to loss of importance information, as fewer combinations of data instances are investigated.
Here, we show how these limitations can be alleviated using a fuzzy logic approach for representation of importance. This accounts for multiple levels of uncertainty, whilst simultaneously simplifying interpretation of results. Fuzzy logic defines data/context-dependent intervals of importance that make it easier to identify data regions where specified features have high or low importance for decision making. This level of interpretation granularity will accelerate data-driven intelligent research and decisions by identifying critical data instances that enhance prognosis and health management, manufacturing, medicine and design of new materials.

\section{Fuzzy Ensemble Feature Importance}
\label{FEFI}

\begin{figure*}[!ht]
\centering
\scalebox{0.65}
{
\includegraphics{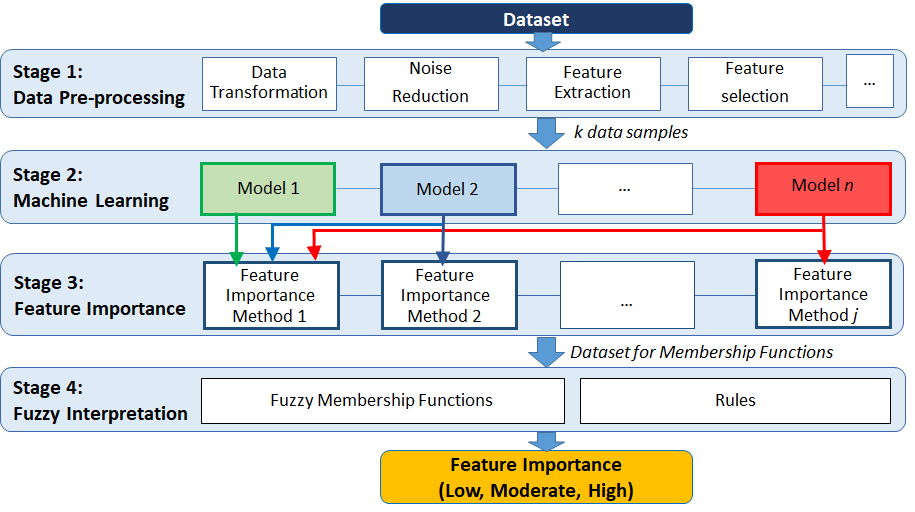}
}
\caption{The fuzzy feature importance framework. }
\label{fig:fuzzysystem}
\end{figure*}

\begin{figure*}[!ht]
\centering
\scalebox{0.6}
{
\includegraphics{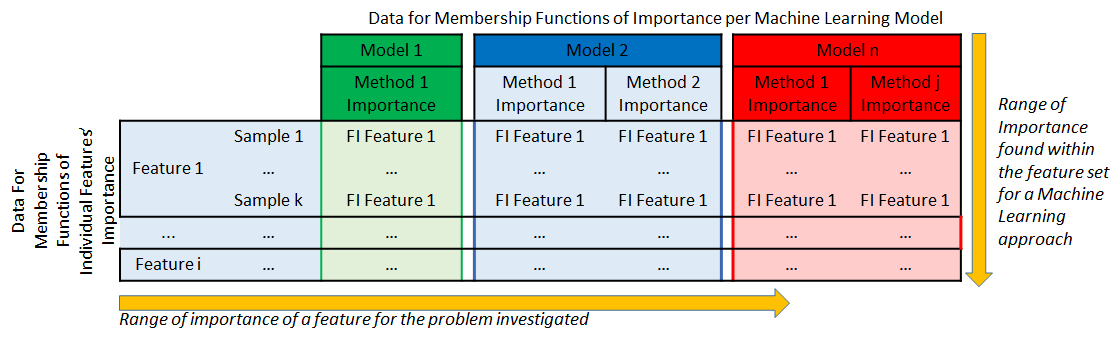}
}
\caption{Data from Stage 3}
\label{fig:DataFromStage3}
\end{figure*}


Our approach uses an ensemble of ML models coupled with multiple FI techniques to generate a large dataset of FI values. This dataset is analysed by a fuzzy logic (FL) system that specifies, for each ML and for each feature, low, moderate, or high importance. A flowchart of our framework is shown in Figure~\ref{fig:fuzzysystem} and the four stages in the framework are described in detail in the following sections.



\subsection{Stage 1: Data Pre-processing}

The data is pre-processed to manage missing values and outliers, identify relevant features, and to perform data normalisation. Feature reduction, engineering, or selection can be used to remove redundant, low variance, and less relevant features, reducing noise in the data. The pre-processed data are subsequently transformed to a format suitable for training the ML models.

The pre-processed data is partitioned into k equally sized subsets for model training using k-fold cross validation~\cite{berrar2019cross} or any other type of bootstrapping. In each training step, 1 subset is held out and the remaining k-1 subsets are used to train the model. The training process is repeated k times, holding out a different subset each time. The k models that result allows the variance in feature gradients and feature contributions to be calculated.
 
\subsection{Stage 2: Machine Learning Predictions}
An ensemble of models is generated using n different ML algorithms, each algorithm being trained \textit{k} times on the partitioned subsets of data. It is important to note that the framework can be extended to include any type of ML method and model-agnostic feature importance technique.  This produces \textit{k * n} trained models. We optimise hyperparameters to ensure that the most accurate models are obtained for the specified problem.

\subsection{Stage 3: Feature Importance}
Each trained model employs a subset of j FI techniques and a validation data subset to produce a large set of FI coefficients for each feature and importance method (see  Figure~\ref{fig:DataFromStage3}). The dataset of FI coefficients encompasses the variance in the learning characteristics and performance of ML approaches, the variance in calculating importance by the various FI techniques, and the variance of feature contributions in the data.


\begin{figure*}[!ht]
\centering
\scalebox{0.5}{
\includegraphics{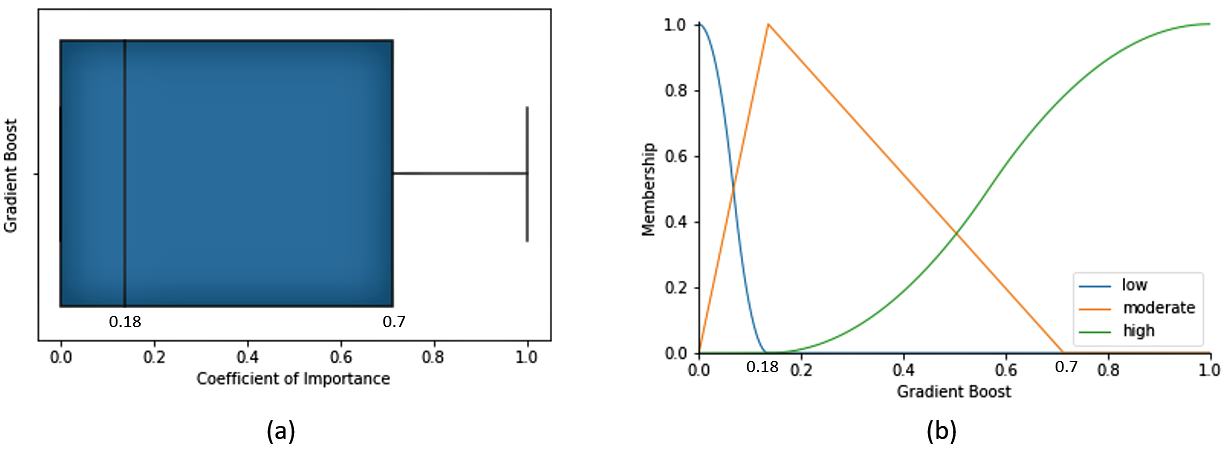}
}
\caption{Membership functions generated from boxplot distribution: a) boxplot distribution of important values, and b) extracted MFs using our boxplot data-driven approach.}
\label{fig:boxplotMF}
\end{figure*}

\subsection{Stage 4: Fuzzy Interpretation}
Here, we generate MFs and inference rules from the data obtained from stage 3. These elucidate whether a feature is of low, moderate, or high importance for each ML method used and how important a feature is after fusing importances from the different approaches. 

\subsubsection{Stage 4 (a); Generating Membership Functions}
\label{generate MFs}
The data set of FI coefficients are normalised to the same scale to ensure unbiased and consistent representation of importance. ML MFs are generated using the FI coefficients from each ML method  (i.e. each `Model' column in Figure~\ref{fig:DataFromStage3} will produce a MF), representing the range of importance in the feature set. Feature MFs are generated by combining the ML MFs for each feature (each `Feature’ row and each `Model’ column will produce MFs) using the minimum (min) and maximum (max) operators. The min operator combines the lower bounds of the MFs and the max operator combines the upper bounds. The feature MFs represent how important a feature is relative to other features for each ML method. The output MF (final importance MF) is generated using the actual (ground truth) FI coefficients for the entire feature set. The output MF presents the final importance of features after combining their importance from the different ML approaches. The MFs (ML, feature and output MFs) consist of the linguistic fuzzy sets (FSs) `low’, `moderate’ or `high’ importance obtained from boxplot distributions of the FI coefficients as proposed by Mase \textit{et al.}~\cite{mase2020capturing}. The five class summary (minimum (min), first quartile (Q$_1$), median (Q$_2$), third quartile (Q$_3$)  and maximum (max)) from the boxplot distribution are used to construct the MFs. We use Z and S-MFs for `low' and `high' FSs as they are most suitable for representing the extreme boundaries with maximum degrees or likelihood of membership. For instance, we expect a feature's importance coefficient of 0 to be in the `low' FS with a degree of 1 (maximum likelihood) and an importance coefficient of 1 to be in the `high' FS with a degree of 1. The Z-MF is generated using the minimum and median values of the distribution and the S-MF employs the median and maximum values. For `moderate' FS, we use Triangular MFs as the most efficient way to represent intermediate FSs for our problem. We use first quartile, median and third quartile values to construct triangular-MFs.


Figure~\ref{fig:boxplotMF} shows an example of MFs generated from a boxplot distribution. Fig.~\ref{fig:boxplotMF}(a) represents the distribution of importance coefficients for the Gradient Boost approach after training several models on different data samples and applying suitable FI techniques on the trained models as described in Stages 2 and 3 (Figure~\ref{fig:fuzzysystem}). Using the five-summary values of the boxplot (min=0, Q$_1$=0, Q$_2$=0.18, Q$_3$=0.7 and max=1), we obtain the MFs in Fig.~\ref{fig:boxplotMF}(b). `low' MF = Z(0, 0.18), `moderate' MF = triangular (0, 0.18, 0.7) and `high' MF = S(0.18, 1). The MFs provide a clear representation of the range and variation of importance coefficients in the low, moderate and high categories. For example, the low category in Figure~\ref{fig:boxplotMF}(b) consists of importance coefficients from 0 to 0.18, a lower variation of values compared to moderate and high categories as shown by its narrow support (range).


\subsubsection{Stage 4 (b); Fuzzy rule generation}
After developing the MFs, we employ the Wang-Mendel~\cite{wangMendel} approach to learn  rules that map the FI coefficients from the different ML approaches (inputs) to the final description of importance (output). The Wang-Mendel is most applicable for ground truth data as it uses a supervised learning approach to capture the appropriate mappings between the inputs and output. Using the data structure in Fig.~\ref{fig:DataFromStage3}, each feature \textit{i} has a ground truth importance coefficient, and each feature sample \textit{k} has a set of FI coefficients obtained by applying multiple (ML, FI) pairs. For example, feature 1 - sample 1 in Fig.~\ref{fig:DataFromStage3} will produce the following input-output pairs:

\begin{equation}
\centering
\begin{split}
       & (FI^{1,1}_{ML_1},\quad FI^{1,1}_{ML_2}, \quad FI^{1,1}_{ML_n};\quad y^1),\\\\ 
       & \quad  \quad  \quad  \quad \quad (FI^{1,2}_{ML_2}, \quad FI^{1,2}_{ML_n}; \quad y^1),\\
       & \quad \quad \quad \vdots   \quad  \quad   \quad  \quad   \quad  \quad   \quad  \quad   \quad  \vdots\\
       &  \quad  \quad   \quad  \quad   \quad \quad   \quad \quad \quad (FI^{1,j}_{ML_n}; \quad y^1),
\end{split}
\end{equation}
where ($FI^{1,1}_{ML_1}$) denotes the corresponding importance coefficient of Feature $1$ obtained from ML model $1$ and FI method $1$ and $y^1$ denotes the ground truth importance coefficient for Feature $1$. Each input-output pair is assigned to the corresponding ML and output MFs to produce the rules and the number of occurrences of each rule (weight). The rule weights are utilised to implement a rule reduction procedure for conflicting rules. For example, Table~\ref{tab:fuzzy_rules} shows sample rules for the data in Fig.~\ref{fig:DataFromStage3}. We observe that rule 2 and rule 5 are in conflict, with weights 5 and 2 respectively. We eliminate rule 5 as it has a smaller weight. Other rule reduction procedures, such as evidence combinational models~\cite{alvarez8proposal} may be applied in this stage. The rules could be further improved by domain experts due to their simple linguistic representation. For a more elaborate description of the Wang-Mendel rule generation method, please refer \cite{wangMendel,alvarez8proposal}.

\begin{table*}[htbp!]
\centering
\caption{Sample rules mapping FI coefficients obtained from multiple ML models and FI techniques to the final FI description)}
\label{tab:fuzzy_rules}
\scalebox{1.0}{
\begin{tabular}{|c|c|c|}
\hline
 \textbf{Rule No} & \textbf{Rule Description} & \textbf{Weight}\\
\hline
         1 &   \textbf{IF} $ML_1$ \textbf{is} 'low'  \textbf{and} $ML_2$ \textbf{is} 'low'  \textbf{and} $ML_n$ \textbf{is} 'low'   \textbf{THEN} Output \textbf{is} 'low'  & 5\\
\hline
         2 &   \textbf{IF} $ML_1$ \textbf{is} 'high'  \textbf{and} $ML_2$ \textbf{is} 'high'  \textbf{and} $ML_n$ \textbf{is} 'high'  \textbf{THEN} Output \textbf{is} 'high' & 5 \\
\hline
         3 &   \textbf{IF} $ML_1$ \textbf{is} 'moderate'  \textbf{and} $ML_2$ \textbf{is} 'moderate'  \textbf{and} \textbf{is} $ML_n$ 'low'  \textbf{THEN} Output \textbf{is} 'low' & 4 \\
\hline
         4 &    \textbf{IF} $ML_1$ \textbf{is} 'moderate'  \textbf{and} $ML_2$ \textbf{is} 'moderate'  \textbf{and} $ML_n$ \textbf{is} 'high'  \textbf{THEN} Output \textbf{is} 'high' &  4 \\         
\hline
         5 &   \textbf{IF} $ML_1$ \textbf{is} 'high'  \textbf{and} $ML_2$ \textbf{is} 'high'  \textbf{and} $ML_n$ \textbf{is} 'high'  \textbf{THEN} Output \textbf{is} 'moderate' & 2   \\
\hline

\end{tabular}}
\end{table*}

\subsubsection{Stage 4 (c); Inference}

The rules utilised during inference fuse the importance of a feature from multiple ML approaches to obtain a final importance (low, moderate, high). To achieve this, we employ the Mamdani inference technique~\cite{mamdani1975experiment}, a rule-based fuzzy inference method. The technique converts FI coefficients into FSs (fuzzification) using the MFs generated in Stage 4 (a), computes rule strengths, and uses these to estimate feature importance as low, moderate, or high. The rule strengths are computed using  `AND' (minimum of membership degrees), `OR' (maximum of membership degrees) and `NOT' (1 minus membership degree)~\cite{zadeh1975fuzzy}. We use the following sample rule to illustrate the calculation of rule strengths and final FI importance:

\textit{ Rule 1: \textbf{IF} $ML_1$ is 'low' \textbf{AND} $ML_2$ is 'low' \textbf{AND} $ML_n$is 'low' \textbf{THEN} Output is 'low'}

\begin{figure*}[!ht]
\centering
\scalebox{0.5}{
\includegraphics{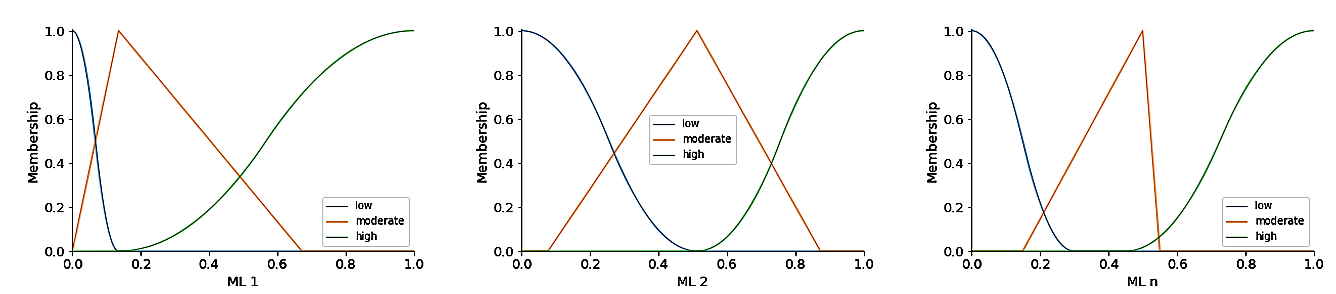}
}
\caption{Sample membership functions for FI coefficients produced by ML approaches.}
\label{fig:MF_ML}
\end{figure*}

\begin{figure*}[!ht]
\centering
\scalebox{0.5}{
\includegraphics{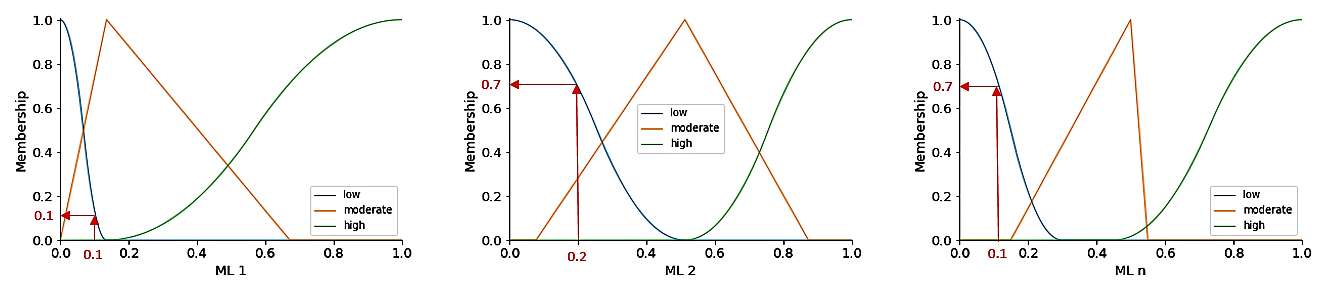}
}
\caption{Fuzzification of FI coefficients using Rule 1 and ML membership functions in Fig.~\ref{fig:MF_ML}.}
\label{fig:FuzzifyFI}
\end{figure*}

\noindent
Assume $ML_1$, $ML_2$, $ML_n$ have the MFs found in Figure~\ref{fig:MF_ML} and the FI coefficients produced for feature 1 by ${ML_1}$ is 0.1 , ${ML_2}$ is 0.2 , and ${ML_n}$ is 0.1 using FI Method 1. The fuzzification process produces the following degrees of membership for the FSs in Rule 1 (as shown in Figure~\ref{fig:FuzzifyFI}):

\begin{equation*}
\centering
\begin{split}
ML_1(low) \quad for \quad 0.1 \quad  = \quad 0.1 \\
ML_2(low) \quad for \quad 0.2 \quad  = \quad 0.7 \\
ML_n(low) \quad for \quad 0.1 \quad  = \quad 0.7 \\
\end{split}
\end{equation*}


\noindent
Using the rule operators in Rule 1 (AND), we obtain the rule strength = $ML_1$(low) AND $ML_2$(low) AND $ML_n$ = min[0.1, 0.7, 0.7] = 0.1

The final step in the Mamdani inference method is to aggregate rule strengths of the entire rule set produced by the rule generation method. This is achieved by finding the maximum rule strength~\cite{mamdani1975experiment} for rules with similar output FSs (e.g. the `THEN' part of rules). The aggregated rule strengths represent the likelihood of the final FI belonging a specific category. For example, if the data from Stage 3 of the framework produce six rules with the following rule strengths and output FS for the FI coefficients of feature 1: 

\begin{equation*}
\begin{split}
& Rule 1:Strength = 0.1,\quad Output=low\\
& Rule 2:Strength = 0 ,\quad Output=low\\
& Rule 3:Strength = 0.2,\quad Output=moderate\\
& Rule 4:Strength = 0.1,\quad Output=moderate\\
& Rule 5:Strength = 0.5,\quad Output=high\\
& Rule 6:Strength = 0.8,\quad Output=high\\
\end{split}
\end{equation*}
\noindent
The final FI of feature 1 using the above rules is: 

\begin{itemize}
\item max(0.1,0)= 10\% likelihood of low importance\\
\item max(0.2,0.1)= 20\% likelihood of moderate importance\\
\item max(0.5,0.6)= 60\% likelihood of high importance\\
\end{itemize}

\section{Experimental design}
\subsection{Data generation}
\begin{figure*}[!htpb]
\centering
\scalebox{0.56}{
\includegraphics{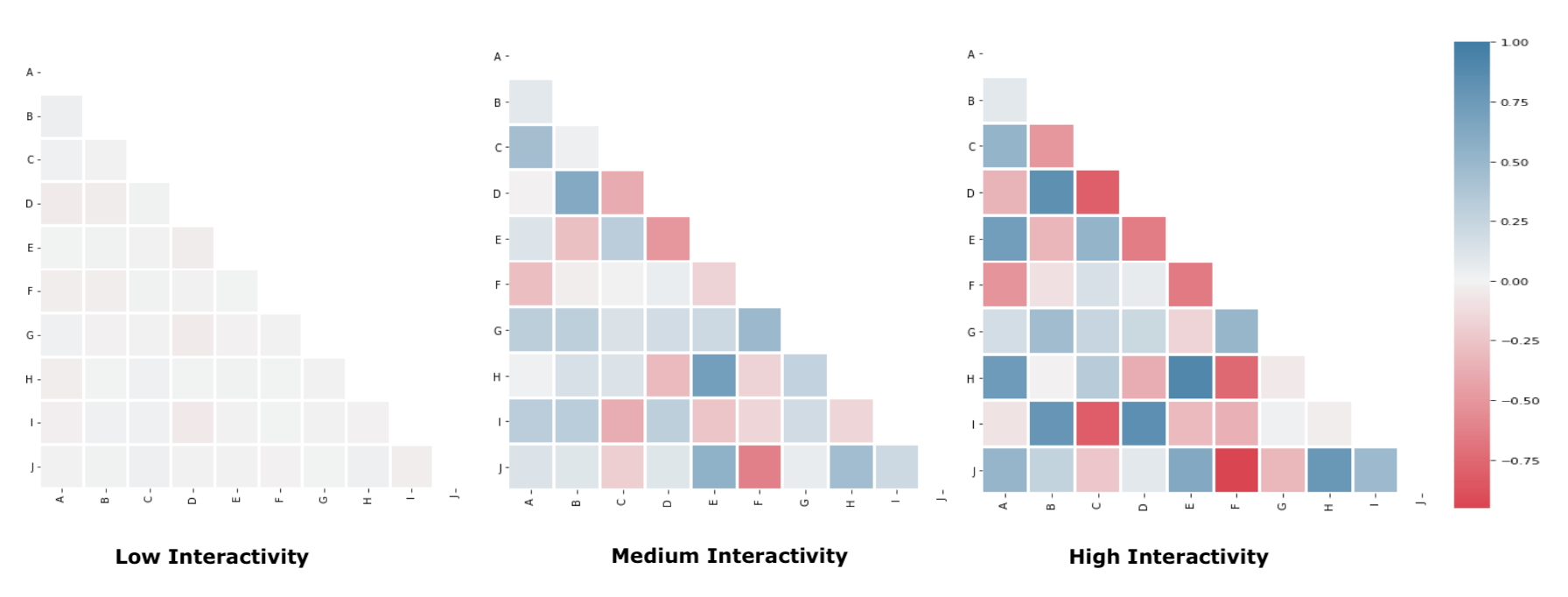}
}
\caption{Pearson correlation matrix for datasets with different interaction strength between ten synthetic features investigated, labelled from A to J.}

\label{fig:corr_level}
\end{figure*}

We validate our approach using synthetic data described by Rengasamy \textit{et al.}~\cite{rengasamy2020towards}, as it allows us to verify the efficacy of feature importance produced by FEFI against the ground truth. Nine datasets are used for regression, with 2000 instances and different number of features, feature informative levels, feature noise levels, or feature interaction levels. Feature informative levels affects the percentage of features contributing to the output data. For example, for ten features with 50\% feature informative levels indicates that five features affects the data output while the remaining five features is independent of the output. Subsequently, the noise levels in features are adjusted using the Gaussian noise standard deviation. Larger values push feature values further from the average, decreasing $Signal/Noise$ ratios. Feature interaction strengths are an additional  characteristic not investigated by Rengasamy \textit{et al.} Feature interaction levels of data determines the strength of interaction between features. It can be varied by changing the effective rank of the matrix, i.e., the maximum number of linearly independent features in a matrix. If a matrix has three columns (features) and the first two columns are linearly independent, the matrix has a rank of at least 2. The feature matrix ranks are generated by the \texttt{make\_regression} function in Python \textit{scikit-learn} library~\cite{scikit-learn}. Figure~\ref{fig:corr_level} shows the correlation matrix for the three datasets with fully independent features compared to those with highly correlated features. The correlation between features is calculated using Pearson correlation.

\begin{table}[H]
\centering
\caption{Datasets with different feature interaction, number of features, features informative level, and noise's standard deviation (std) tested on the proposed Fuzzy Ensemble Feature Importance methods.}
\label{tab:datasets}
\begin{tabular}{|l|c|c|c|c|}
  \hline
  \textbf{Dataset}   & \textbf{Number of} & \textbf{Feature} & \textbf{Feature} & \textbf{Feature}\\
  \textbf{Number}   & \textbf{features} & \textbf{interaction} & \textbf{informative level} & \textbf{noise std}\\
  \hline
     1 & \cellcolor{blue!25}10 & Low & 90\% & 0.5\\
  \hline
     2 &\cellcolor{blue!25} 30 & Low & 90\% & 0.5 \\
  \hline
     3 & \cellcolor{blue!25}50 & Low & 90\% & 0.5 \\
  \hline
       4 & 10 & \cellcolor{red!25}Moderate & 90\% & 0.5\\
  \hline
       5 & 10 & \cellcolor{red!25}High & 90\% & 0.5\\
  \hline
       6 & 10 & Low & \cellcolor{cyan!25}20\% & 0.5\\
  \hline
       7 & 10 & Low & \cellcolor{cyan!25}50\% & 0.5\\
  \hline
       8 & 10 & Low & 90\% &  \cellcolor{green!25}2.0\\
  \hline
       9 & 10 & Low & 90\% & \cellcolor{green!25}5.0\\
  \hline
\end{tabular}

\end{table}

\begin{figure}[!htpb]
\centering
\scalebox{0.25}{
\includegraphics{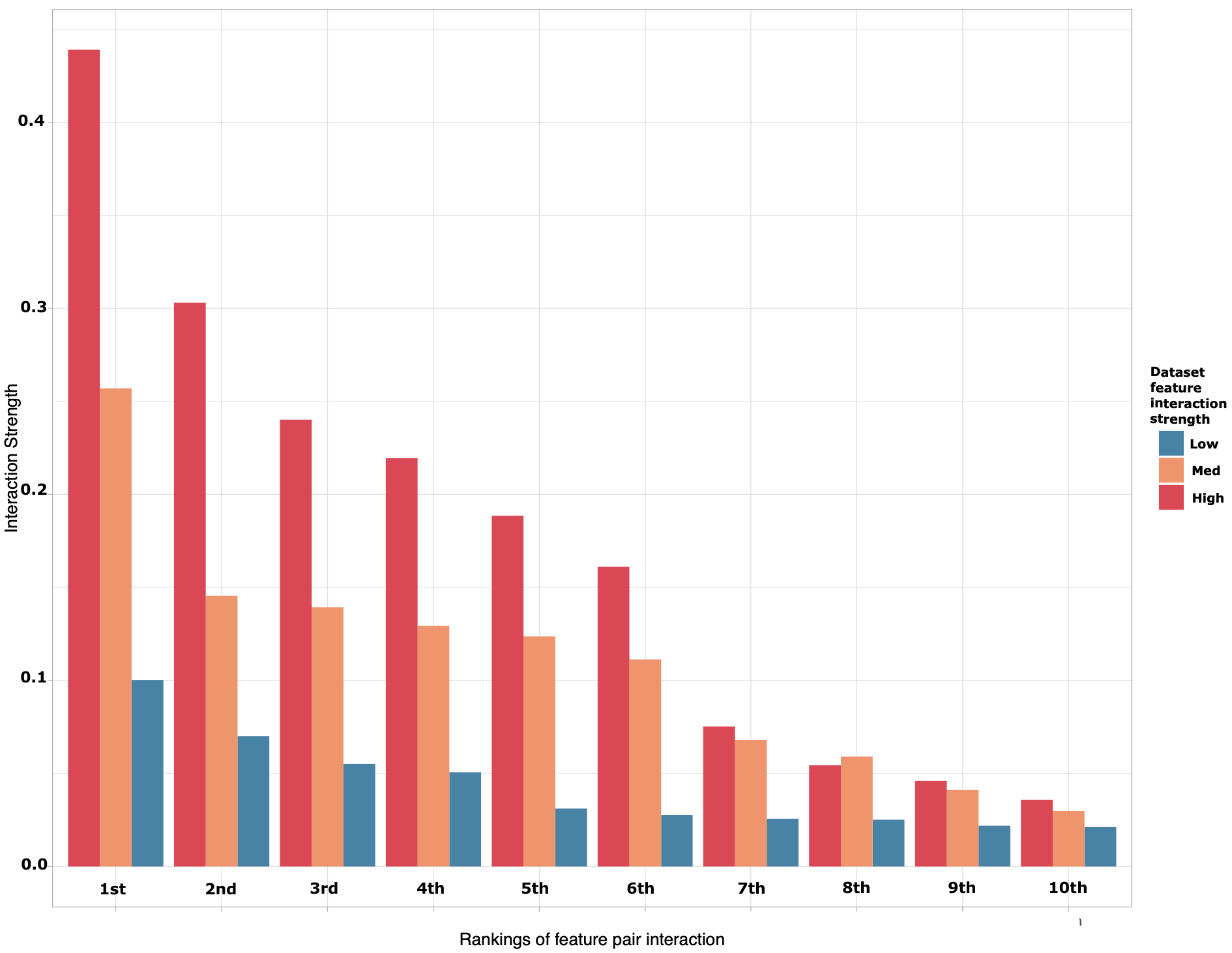}
}
\caption{Interaction strength between pairs of features for each of the three datasets. The x-axis denotes the feature pairs with the top 10 highest interaction from each of the datasets.}
\label{fig:interaction_level}
\end{figure}

Additionally, different feature interaction intensities are added to each dataset, as illustrated in Figure~\ref{fig:interaction_level}. Feature interaction occurs when a feature affects one or more features, making it difficult to discern its actual FI. The interaction effect between pairs of features is calculated using Friedman's H statistics~\cite{friedman2008predictive}.  Table~\ref{tab:data_params} shows summary of different datasets tested in this paper. The dataset's properties are altered one at a time.

\begin{figure*}[!htpb]
\centering
\scalebox{0.56}{
\includegraphics{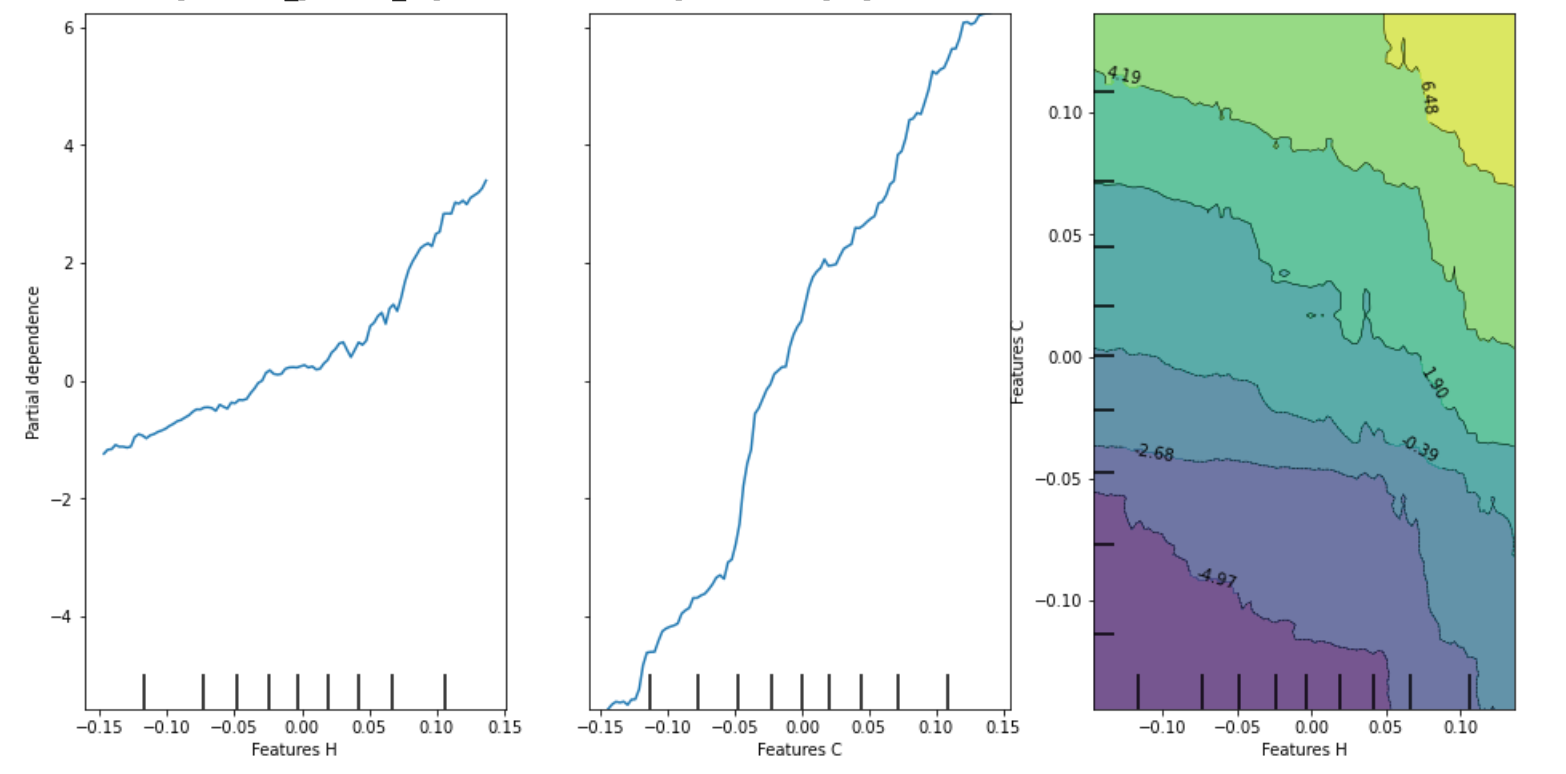}
}
\caption{The partial dependence plot of two features on the output. The left and the center plot shows the respective partial dependence of the respective feature to output. The right plot shows the combined response on the output for the two features.}

\label{fig:pdp}
\end{figure*}

We use partial dependence plots (PDPs) to analyse and illustrate the relationship between the features and output. It shows that for high feature interaction strengths the output response surface may not be linear. The PDPs show the marginal influence of one or two combined features on the output using Monte Carlo method across the dataset~\cite{friedman2001greedy}. The PDP in Figure~\ref{fig:pdp} illustrates the non-linear relationship between two features and the output in a high interactive region of the dataset.

\begin{table}[!htp]
\centering
\caption{Datasets with different feature interaction intensities, number of features, features informative level, and noise's standard deviation tested in the FEFI framework.}
\label{tab:data_params}
\begin{tabular}{|l|c|}
  \hline
  \textbf{Dataset properties}   & \textbf{Values of properties} \\
  \hline
  Number of features & 10, 30, 50 \\
  \hline
  Feature interaction  & Low, Medium, High \\
  \hline
  Features informative level & 20\%, 50\%, 90\% \\
  \hline
  Noise's standard deviation & 0.5, 2.0, 5.0 \\
  \hline
\end{tabular}

\end{table}

\subsection{Machine learning methods selection and configurations}
Five ML methods were implemented: random forest (RF), gradient boost (GB), extra trees (ET), support vector regression (SVR), multiLayer perceptron (MLP). Three FI methods: PI, SHAP, and Gini Importance are used. Diverse ML and FI techniques are chosen to show the benefit of FEFI in modelling the uncertainties in FI due to differences in learning and FI identification characteristics. The hyperparameters for each ML method are shown in Table~\ref{tab:model-hyper}.

\begin{table}[!htp]
\centering
\caption{Hyperparameters for each of the machine learning approaches employed to test the Fuzzy Ensemble Feature Importance (FEFI) framework}
\label{tab:model-hyper}
\scalebox{0.8}{
\begin{tabular}{|l|l|c|} 
\hline
\multicolumn{1}{|c|}{ML Approaches}                 & \multicolumn{1}{c|}{Hyperparameters} & Values         \\ 
\hline
\hline
\multirow{6}{*}{Gradient Boosting Regressor} & Loss                                 & Least squares  \\ 
\cline{2-3}
                                             & Learning rate                        & 0.1            \\ 
\cline{2-3}
                                             & Number of boosting stages            & 50             \\ 
\cline{2-3}
                                             & Splitting criterion                  & Friedman MSE   \\ 
\cline{2-3}
                                             & Minimum samples to split             & 2              \\ 
\cline{2-3}
                                             & Maximum depth                        & 3              \\ 
\hline
\hline
\multirow{5}{*}{Random Forest Regressor}     & Number of Trees                      & 50             \\ 
\cline{2-3}
                                             & Max depth                            & None           \\ 
\cline{2-3}
                                             & Splitting Criterion                  & MSE            \\ 
\cline{2-3}
                                             & Minimum samples to split             & 2              \\ 
\cline{2-3}
                                             & Bootstrap                            & True           \\ 
\hline
\hline
\multirow{5}{*}{Extra Trees Regressor}       & Number of Trees                      & 50             \\ 
\cline{2-3}
                                             & Max depth                            & None           \\ 
\cline{2-3}
                                             & Splitting Criterion                  & MSE            \\ 
\cline{2-3}
                                             & Minimum samples to split             & 2              \\ 
\cline{2-3}
                                             & Bootstrap                            & False          \\ 
\hline
\hline
\multirow{4}{*}{Support Vector Regressor}    & Kernel                               & Linear        \\ 
\cline{2-3}
                                             & Tolerance                            & 0.001          \\ 
\cline{2-3}
                                             & Regularisation                       & 1.0            \\ 
\cline{2-3}
                                             & Epsilon                              & 0.1            \\
\hline
\hline
\multirow{5}{*}{MultiLayer Perceptron}       & Hidden Layer Size                      & 50             \\ 
\cline{2-3}
                                             & Activation function                            & ReLU           \\ 
\cline{2-3}
                                             & Optimiser                  & Adam            \\ 
\cline{2-3}
                                             & L2 Regulariser             & 0.0001              \\ 
\cline{2-3}
                                             & Learning Rate                            & 0.001          \\ 
\hline
\end{tabular}}
\end{table}

\subsection{Evaluation metrics}

The FEFI approach is evaluated and compared to mean and majority vote EFI using 4 different data properties: features interaction, number of features, features informative level, and noise level of features. The mean absolute error (MAE) and root mean squared error (RMSE) metrics compared these properties to the ground truth feature importance. Three of the four data properties, except feature interaction were tested  by Rengasamy \textit{et al.}~\cite{rengasamy2020towards}. Each data property was further split into 3 different subsets: training; test; and whole (train + test) set for feature importance quantification and comparisons between mean EFI, majority vote EFI, and FEFI.

\section{Results and Discussion}
\subsection{Results}
Here we compare the results of FEFI to the best crisp fusion strategies from Rengasamy \textit{et al}, the mean and majority vote across the different datasets shown in the Experimental design section (Table~\ref{tab:datasets}).

Table~\ref{tab:main_results_features} shows the errors in calculating the final FI using the different fusion strategies as the number of features in the data increases (datasets 1, 2 and 3 in Table~\ref{tab:datasets}). Increasing the number of data features with a 90\% feature informative level made it more difficult for crisp strategies to quantify feature importance. This is illustrated by the significant increase in FI errors from 10 to 30 features [p-value $<$ 0.001 for all data subsets] and from 10 to 50 features [p-value $<$ 0.001 for all data subsets]. This is due to the inefficiency of crisp methods in capturing the increased variation of FI coefficients produced by additional features. In contrast, FEFI exhibited no significant differences in FI errors for different number of features [p-value $>$ 0.1 for all data subsets] and significantly lower FI errors with 30 and 50 data features compared to mean and majority vote [p-value $<$ 0.001 for all data subsets]. This indicates that FEFI efficiently captures the increased FI uncertainty caused by extra number of features compared to the crisp strategies.

\begin{table}[!htp]
\centering
\caption{MAE and RMSE comparison between three different frameworks for feature importance fusion/ensemble: (1) Mean, (2) Majority Vote, (3) FEFI on three datasets with different number of features using three data subsets}
\label{tab:main_results_features}
\resizebox{\columnwidth}{!}{
\begin{tabular}{|c|l|l|l|l|l|l|l|} 
\hline
\multirow{2}{*}{\begin{tabular}[c]{@{}c@{}}Number of\\~features \end{tabular}} &                      & \multicolumn{2}{c|}{Mean}                   & \multicolumn{2}{l|}{Majority Vote}          & \multicolumn{2}{c|}{FEFI}                   \\ 
\cline{2-8} & Data subset  & MAE  & RMSE   & MAE   & RMSE & MAE  & RMSE                  \\ 
\hline
\multirow{3}{*}{\begin{tabular}[c]{@{}c@{}}Dataset 1:\\~10 \end{tabular}}  & Whole  & 0.146   & 0.168   & \textbf{0.124}   & \textbf{0.143} & 0.148   & 0.181  \\
\cline{2-8}
              & Train   & 0.146   & 0.168   &\textbf{0.124 }  & \textbf{0.143 }  & 0.126   & 0.156  \\ 
\cline{2-8}
              & Test  & 0.154    & 0.181  & \textbf{0.128 }  & \textbf{0.153}    & 0.135  & 0.169   \\ 
\hline
\hline
\multirow{3}{*}{\begin{tabular}[c]{@{}c@{}}Dataset 2:\\~30 \end{tabular}} & Whole   & 0.379  & 0.435 &  0.397  & 0.454 & \textbf{0.107} & \textbf{0.133} \\ 
\cline{2-8}
         & Train   & 0.379  & 0.434  &  0.397  &  0.454  & \textbf{0.109}  & \textbf{0.135}  \\ 
\cline{2-8}
          & Test  & 0.377  & 0.433  &  0.398   & 0.455 & \textbf{0.117}  & \textbf{0.148}  \\ 
\hline
\hline
\multirow{3}{*}{\begin{tabular}[c]{@{}c@{}}Dataset 3:\\~50 \end{tabular}}   & Whole   & 0.297  & 0.365  &  0.293 & 0.360 &  \textbf{0.128}  & \textbf{0.155}  \\ 
\cline{2-8}
         & Train  &  0.298 & 0.365  & 0.294  & 0.361 & \textbf{0.131}  &  \textbf{0.155} \\ 
\cline{2-8}
            & Test  & 0.286   & 0.354  & 0.284  &  0.351  & \textbf{0.128}  &  \textbf{0.155}  \\ 
\hline
\end{tabular}}

\end{table}

Table~\ref{tab:main_results_interaction} shows errors in calculating final FI as the interactions between features in the data increase (datasets 1, 4 and 5 in Table~\ref{tab:datasets}). As with the number of features, there is a significant increase in errors for mean and majority vote fusion strategies when the interactions between data features increase from low to moderate [p-value$<$0.05 for all data subsets] and from low to high interaction [p-value$<$0.02 for all data subsets]. However, there is no significant increase in errors from medium to high feature interaction [p$>$0.05 for all data subsets]. For FEFI, we again observe no significant difference in FI errors for different levels of feature interaction [p$>$0.1 for all data subsets] and significantly lowers FI errors for medium and high feature interaction levels compared to mean and majority vote [p$<$0.001 for all data subsets]. Again, this indicates that the fuzzy approach efficiently captures the increased variation of FI coefficients caused by higher feature interactions compared to the crisp strategies.

\begin{table}[!htp]
\centering
\caption{MAE and RMSE comparison between three different frameworks for feature importance fusion/ensemble: (1) Mean, (2) Majority Vote, (3) FEFI on three different features interaction level using three data subsets.}
\label{tab:main_results_interaction}
\resizebox{\columnwidth}{!}{
\begin{tabular}{|c|l|l|l|l|l|l|l|} 
\hline
\multirow{2}{*}{\begin{tabular}[c]{@{}c@{}}Interaction\\~level \end{tabular}} &                      & \multicolumn{2}{c|}{Mean}                   & \multicolumn{2}{l|}{Majority Vote}          & \multicolumn{2}{c|}{FEFI}                   \\ 
\cline{2-8} & Data subset  & MAE  & RMSE   & MAE   & RMSE & MAE  & RMSE                  \\ 
\hline
\multirow{3}{*}{\begin{tabular}[c]{@{}c@{}}Dataset 1:\\~Low \end{tabular}}  & Whole  & 0.146   & 0.168   & \textbf{0.124}   & \textbf{0.143} & 0.148   & 0.181  \\
\cline{2-8}
              & Train   & 0.146   & 0.168   & \textbf{0.124}   & \textbf{0.143}   & 0.126   & 0.156  \\ 
\cline{2-8}
              & Test  & 0.154    & 0.181  & \textbf{0.128}   & \textbf{0.153}    & 0.135  & 0.169   \\  
\hline
\hline
\multirow{3}{*}{\begin{tabular}[c]{@{}c@{}}Dataset 4:\\~Medium \end{tabular}}    & Whole   & 0.203   & 0.248 & 0.220  & 0.275  & \textbf{0.141}  & \textbf{0.189}  \\ 
\cline{2-8}
         & Train   & 0.203  & 0.249  & 0.221   & 0.276   & \textbf{0.135}  & \textbf{0.179}  \\ 
\cline{2-8}
          & Test  & 0.202  & 0.251  & 0.216   & 0.273 & \textbf{0.148}  & \textbf{0.188}  \\ 
\hline
\hline
\multirow{3}{*}{\begin{tabular}[c]{@{}c@{}}Dataset 5:\\~High \end{tabular}}  & Whole   & 0.222  & 0.288  & 0.249  & 0.314 & \textbf{0.125}   & \textbf{0.164}  \\ 
\cline{2-8}
         & Train  & 0.222  & 0.288  & 0.250  & 0.315  & \textbf{0.139}  & \textbf{0.177}   \\ 
\cline{2-8}
            & Test  & 0.223  & 0.291   & 0.226  & 0.288   & \textbf{0.144}   & \textbf{0.180}                 \\ 
\hline
\end{tabular}}

\end{table}

Table~\ref{tab:main_results_inform} presents the FI errors obtained from the different fusion strategies as the informative level of features increases i.e. datasets 1, 6 and 7 in Table~\ref{tab:datasets}. We observe a significant rise in FI errors for mean and majority vote ensembles as feature informative levels increase from 20\% to 50\% [p$<$0.001 for all data subsets], 20\% to 90\% [p$<$0.0001 for all data subsets] and 50\% to 90\%[p$<$0.005 for all data subsets]. This means the variation between  calculated FI coefficients increases as the informative level of features increases. However, the results for FEFI show no significant increase in FI errors as feature informative levels increase from 20\% to 50\% [p$>$0.1 for all data subsets], 20\% to 90\% [p$>$0.01 for all data subsets] and 50\% to 90\%[p$>$0.05 for all data subsets]. This implies FEFI shows better performance in capturing increased FI variation produced by an increase in feature informative level.

\begin{table}[!htp]
\centering
\caption{MAE and RMSE comparison between three different frameworks for feature importance fusion/ensemble: (1) Mean, (2) Majority Vote, (3) FEFI on three different features informative level using three different subsets of data.}
\label{tab:main_results_inform}
\resizebox{\columnwidth}{!}{
\begin{tabular}{|c|l|l|l|l|l|l|l|} 
\hline
\multirow{2}{*}{\begin{tabular}[c]{@{}c@{}}Informative\\~level \end{tabular}} &                      & \multicolumn{2}{c|}{Mean}                   & \multicolumn{2}{l|}{Majority Vote}          & \multicolumn{2}{c|}{FEFI}                   \\ 
\cline{2-8} & Data subset  & MAE  & RMSE   & MAE   & RMSE & MAE  & RMSE                  \\ 
\hline
\multirow{3}{*}{\begin{tabular}[c]{@{}c@{}}Dataset 6:\\~20\% \end{tabular}}  & Whole   &  0.008  &  0.011  &   \textbf{0.007}  & \textbf{0.009 }&  0.078  & 0.177  \\ 
\cline{2-8}
              & Train   & 0.008   & 0.011   & \textbf{0.007}  & \textbf{0.009 } &  0.084  &  0.178 \\ 
\cline{2-8}
              & Test   &  0.010   & 0.012  & \textbf{0.008}   & \textbf{0.010}    & 0.078  &  0.178  \\ 
\hline
\hline
\multirow{3}{*}{\begin{tabular}[c]{@{}c@{}}Dataset 7:\\~50\% \end{tabular}}     & Whole   & 0.078  &  0.101 &  \textbf{0.061} &  \textbf{0.086} & 0.111  & 0.164  \\ 
\cline{2-8}
         & Train   & 0.079   & 0.102  & \textbf{0.062}   & \textbf{0.088} & 0.099  &  0.144 \\ 
\cline{2-8}
          & Test  & 0.093  & 0.134  & \textbf{0.078}   & \textbf{0.121} & 0.111  &  0.164 \\ 
\hline
\hline
\multirow{3}{*}{\begin{tabular}[c]{@{}c@{}}Dataset 1:\\~90\% \end{tabular}}  & Whole  & 0.146   & 0.168   & \textbf{0.124}   & \textbf{0.143} & 0.148   & 0.181  \\
\cline{2-8}
              & Train   & 0.146   & 0.168   & \textbf{0.124}   & \textbf{0.143}   & 0.126   & 0.156  \\ 
\cline{2-8}
              & Test  & 0.154    & 0.181  & \textbf{0.128}   & \textbf{0.153}    & 0.135  & 0.169   \\
\hline
\end{tabular}}

\end{table}

Table~\ref{tab:main_results_noise} illustrates the errors obtained from calculating the final FI as the level of noise in data increases from 0.5 noise standard deviation to 2.0 and 5.0 (datasets 1, 8 and 9 in Table~\ref{tab:datasets}). We observe no significant difference in FI errors for mean, majority vote and FEFI fusion strategies as the noise standard deviation increases from 0.5 to 2.0 [p-value$>$0.5 for all data subsets], 0.5 to 5.0 [p$>$0.5 for all data subsets] and 2.0 to 0.5 [p-value$>$0.9 for all data subsets]. This means the variation or uncertainty of FI coefficients is not affected by changes of the level of noise in the data. However, FEFI produces a significantly lower FI error compared to mean and majority vote when the noise standard deviation is 5.0 [p$<$0.05 for whole and test subsets]. This is because of its capability to effectively capture noise in data using its MFs.

\begin{table}[!htp]
\centering
\caption{MAE and RMSE comparison between three different frameworks for feature importance fusion/ensemble: (1) Mean, (2) Majority Vote, (3) FEFI on three different features noise levels using three different subsets of data.}
\label{tab:main_results_noise}
\resizebox{\columnwidth}{!}{
\begin{tabular}{|c|l|l|l|l|l|l|l|} 
\hline
\multirow{2}{*}{\begin{tabular}[c]{@{}c@{}}Noise's Standard\\~Deviation\end{tabular}} &                      & \multicolumn{2}{c|}{Mean}                   & \multicolumn{2}{l|}{Majority Vote}          & \multicolumn{2}{c|}{FEFI}                   \\ 
\cline{2-8} & Data subset  & MAE  & RMSE   & MAE   & RMSE & MAE  & RMSE                  \\ 
\hline
\multirow{3}{*}{\begin{tabular}[c]{@{}c@{}}Dataset 1:\\~0.5 \end{tabular}}   & Whole  & 0.146   & 0.168   & \textbf{0.124}   & \textbf{0.143} & 0.148   & 0.181  \\
\cline{2-8}
              & Train   & 0.146   & 0.168   & \textbf{0.124}   & \textbf{0.143}   & 0.126   & 0.156  \\ 
\cline{2-8}
              & Test  & 0.154    & 0.181  & \textbf{0.128 }  & \textbf{0.153}    & 0.135  & 0.169   \\
\hline
\hline
\multirow{3}{*}{\begin{tabular}[c]{@{}c@{}}Dataset 8:\\~2.0 \end{tabular}}      & Whole   & 0.150 &0.170 & 0.150 & 0.169 & \textbf{0.125}  & \textbf{0.152}  \\ 
\cline{2-8}
         & Train   & 0.150 &0.171 & 0.152  & \textbf{0.169}  & \textbf{0.133}  &  0.180 \\ 
\cline{2-8}
          & Test  & 0.165 & 0.195 & 0.163  & 0.200 & \textbf{0.131}  &  \textbf{0.161} \\ 
\hline
\hline
\multirow{3}{*}{\begin{tabular}[c]{@{}c@{}}Dataset 9:\\~5.0 \end{tabular}}    & Whole   & 0.151 & 0.173 & 0.155 & 0.180 & \textbf{0.117}   & \textbf{0.137}  \\ 
\cline{2-8}
         & Train  & 0.150 & 0.172 & 0.151 & 0.171 & \textbf{0.131}  &  \textbf{0.168}  \\ 
\cline{2-8}
            & Test  & 0.164 & 0.195  & 0.165 & 0.197  &  \textbf{0.117}  &  \textbf{0.137} \\ 
\hline
\end{tabular}}

\end{table}

In summary, the results show that our fuzzy FI fusion approach outperforms mean and majority vote feature importance fusion methods in capturing the variation of FI coefficients caused by increased data dimensionality, complexity and noise. This is because  FEFI explores the data space more thoroughly as it uses multiple samples of data to make decisions about the importance of features. It also uses distributions of the data to provide better definitions of FI and soft boundaries to capture the intermediate uncertainties of FI classification. 

Finally, it is important to mention that majority vote generates apparently lower errors compared to FEFI for 10 features, low feature interaction level, 0.5 noise standard deviation, 20\% , 50\% and 90\% informative levels. However, the difference in errors is not statistically significant [p$>$0.05 for all data subsets] except for 20\% and 50\% feature informative levels where p$<$0.001.

\begin{figure*}[htp]
\centering
\scalebox{0.5}{
\includegraphics{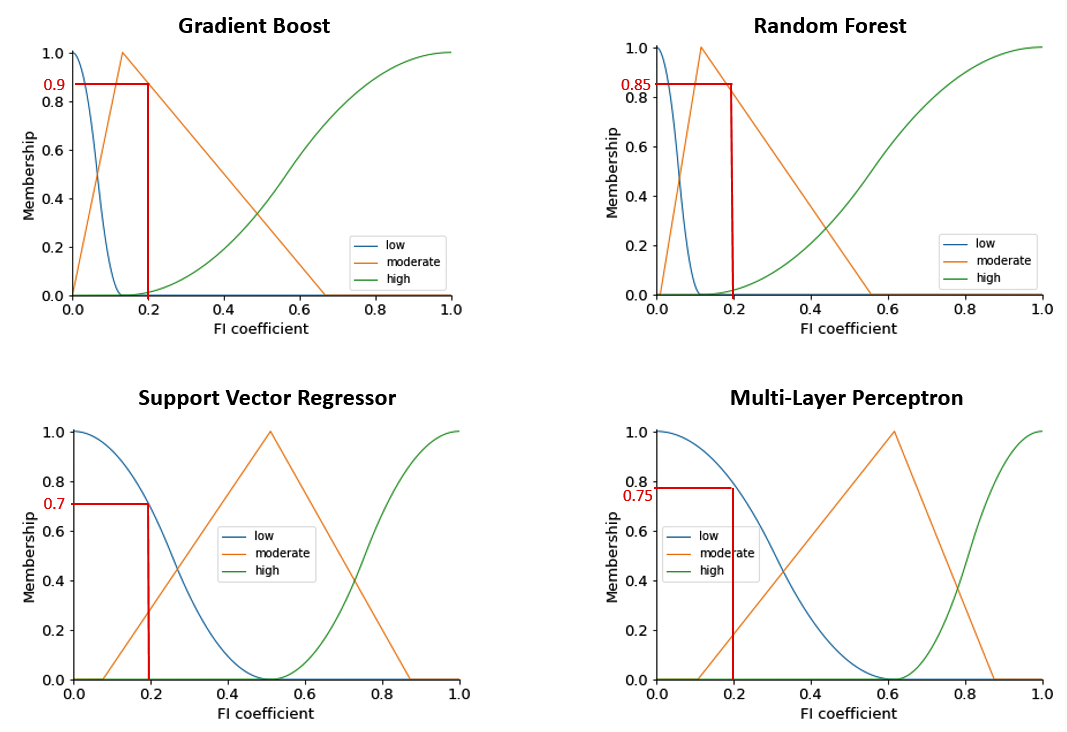}
}
\caption{Membership functions of ML approaches generated from Dataset 1 showing different interpretations of importance for the same FI coefficient}
\label{fig:MFs}
\end{figure*}

\begin{figure}[htp]
\centering
\scalebox{0.45}{
\includegraphics{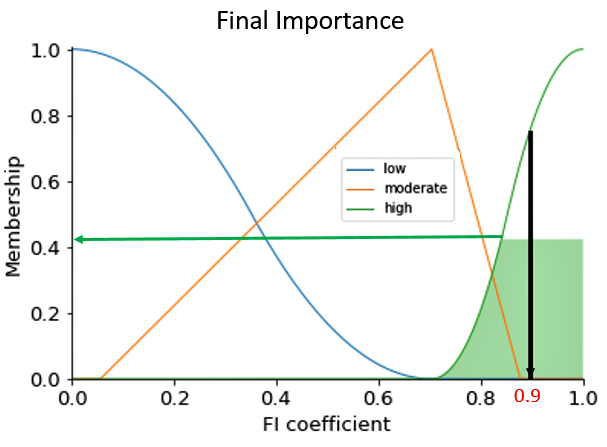}
}
\caption{An example of the final importance of a feature after fusing the importance coefficients obtained from the different ML approaches.}
\label{fig:finaloutput}
\end{figure}

\begin{figure*}[htp]
\centering
\scalebox{0.5}{
\includegraphics{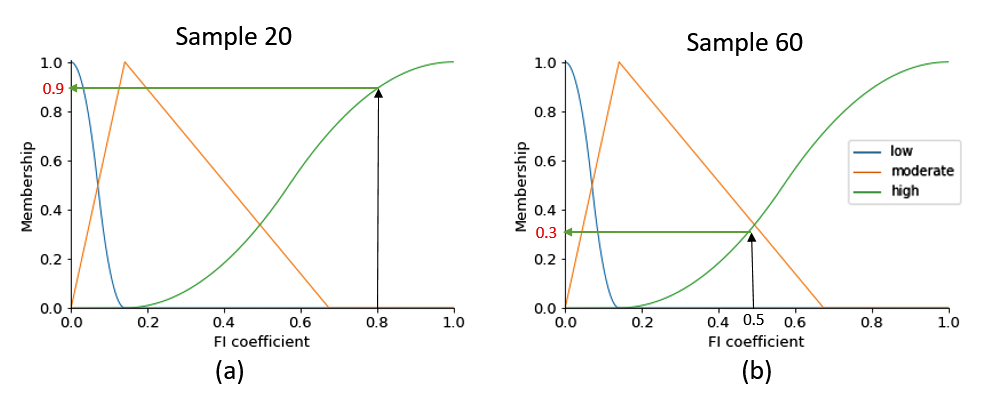}
}
\caption{The same feature having different FI coefficients in different samples of Dataset 1. (a) Sample 20 produces a FI coefficient of 0.8, and (b) Sample 60 produces a FI coefficient of 0.5}
\label{fig:mf_feature}
\end{figure*}

\subsection{Discussion}
Here we discuss the interpretability, advantages and limitations of FEFI in determining importance of features. 

FEFI provides clear and accurate interpretation of feature contribution to different ML models outcomes using the ML MFs.  For example, Fig.~\ref{fig:MFs} shows different interpretations for the same FI coefficient (0.2) by the MFs generated from dataset 1 (refer dataset description in Table~\ref{tab:datasets}) for GB, RF, SVR, and MLP. We observe that a FI coefficient of 0.2 has a 70\% likelihood of low importance for SVR and a 75\% likelihood of low importance for MLP, similar to what we would expect for a coefficient close to 0. However, for GB and RF, a coefficient of 0.2 has a high likelihood of moderate importance and no likelihood of low importance i.e. 90\% likelihood of moderate importance for GB and 85\% for RF. Therefore, without the generation of these MFs, the interpretation of the coefficients produced by the different ML approaches will be misleading. In addition, Figure~\ref{fig:MFs} supports our use of an ensemble of diverse ML approaches to complement outputs. We observe different MF representations for the different ML approaches but similar MFs for approaches with similar learning characteristics. For example, the low importance FS for GB is very narrow compared SVR and MLP but very similar to that of RF due to similar tree-based learning process.

FEFI also provides more reliable interpretations of the final FI for domain experts. Using the output MFs as crisp FI coefficients produced by fusion strategies may be misleading without context.  For example, Figure~\ref{fig:finaloutput} shows the final importance coefficient (0.9) after fusing the importance obtained from the different ML approaches. Without context (the output MF), it might be assumed that 0.9 is a high likelihood of high importance as 0.9 is very close to 1 (maximum likelihood), whereas, we find that 0.9 has a 40\% likelihood of high importance when the entire feature space is considered (indicated by the height of the shaded region in Figure~\ref{fig:finaloutput}). This context provided by FEFI is important in safety-critical and medical diagnosis in providing the extent or critical level of a feature's contribution to the decision.

Furthermore, FEFI shows great potential as a diagnostic tool in identifying data subsets with extreme cases of feature importance (this might be the case of activity cliffs in quantitative structure activity relationship modelling) or with significant variation of feature importance (e.g., for significantly nonlinear response surfaces). This is achieved by producing feature MFs at different data samples or time steps to investigate the importance of a feature in the samples. For example, Fig.~\ref{fig:mf_feature} shows the importance of a feature generated by different data samples of dataset 1 when PI is used to calculate the feature's importance for best performing GB model. We observe that the feature's importance coefficient is different in the samples; 0.8 in sample 20 with a 90\% likelihood of high importance and 0.5 in sample 60 with a 30\% likelihood of high importance. Real-world systems can have more extreme cases as they produce dynamic and multifaceted data where the importance of a feature may vary in different samples due to interactions with other features or supplementary data from other data sources. For example, in healthcare, when multiple sources of data (patient surveys, clinical trials etc) associated with a heterogeneous patient population are merged, features may have different importance depending on patient stratification. 

In real-world systems where the ground truth FI is unavailable, complete automation is limited using FEFI. This is because development of the user-friendly IF/THEN rules requires: interaction with ML experts based on their experience using ML approaches and their expectations of the ML performance on the data; a comprehensive literature review to find out the rules describing the generation of FI by the different ML approaches; or  the use of heuristic rule generation methods such as genetic algorithms~\cite{cordon2001generating}. In addition, the number of rules increases exponentially with an increase in the number of ML approaches, making it difficult for experts to provide rules and reduces the interpretability of the rules.


\section{Conclusions and Future Work}
\label{Conclusions}

This paper introduces a FEFI framework to interpret ensemble of ML methods and FI methods. This is primarily motivated by the uncertainty and unreliable of post-processing FI methods along with the ethical implication of untrustworthy output interpretations. To overcome these issues we proposed a framework that combines ensemble feature importance methods with Fuzzy Logic to utilise the distributions of the FI output to provide better definitions of FI and soft boundaries to capture the intermediate uncertainties of FI classification. Furthermore, Fuzzy Systems have the added advantage of explaining the FI in linguistic term making it easier for non-experts to understand the implication of FI outputs. The combination of robust uncertainty estimations and simple linguistic representation of the FI outputs allows users to make informed decisions --- this is especially important for safety critical applications. We applied the methodology to 9 different synthetic datasets all with varying level of number of features, feature interaction strengths, informative level, and noise level. The usage of synthetic datasets allow us to empirically measure the accuracy of FEFI framework. The results shows that by considering the uncertainty produced in the multi-ML and multi-FI methods settings, FEFI was able to produce more accurate FI outputs compared to mean and majority vote-based ensemble FI frameworks. For future work, we intend to evaluate our methodology using other ML methods such as Neural Networks with more complex learning processes. Additionally, there are multiple ways of capturing uncertainty in data besides Fuzzy Logic, such as Bayesian Statistics. A comparison of the framework using Fuzzy Logic and Bayesian Statistics would provide a greater insight into which uncertainty estimation method is more suitable for providing a more robust and accurate FI interpretation for the end users.
\bibliographystyle{unsrt}
\balance
\bibliography{reference.bib}
\end{document}